\documentclass{article}

\PassOptionsToPackage{numbers, sort&compress}{natbib}

\usepackage{arxiv}

\usepackage{multirow}
\usepackage{graphicx}
\usepackage{enumitem}
\usepackage[utf8]{inputenc} 
\usepackage[T1]{fontenc}    
\usepackage{url}            
\usepackage{booktabs}       
\usepackage{amsfonts}       
\usepackage{nicefrac}       
\usepackage{microtype}      
\usepackage{xcolor}         
\usepackage{array}
\usepackage{adjustbox}
\usepackage{tabularx}
\usepackage{tabulary}
\usepackage[labelfont=bf,textfont=it,font=small]{caption}
\usepackage{hyperref}       
\usepackage{natbib}
\hypersetup{
    colorlinks=true,
    linkcolor={blue},
    citecolor={red},
    urlcolor={blue},
    breaklinks=true,
}
\newcommand{\cref}[2]{\hyperref[#2]{#1~\ref*{#2}}}
\newcommand{\colref}[3]{\hyperref[#2]{#1~\ref*{#2}{#3}}}
\newcommand{\figref}[1]{\cref{Figure}{#1}}

\newcommand{\secref}[1]{\cref{Section}{#1}}
\newcommand{\Eqnref}[1]{\cref{Equation}{#1}}
\newcommand{\tabref}[1]{\cref{Table}{#1}}

\title{ZeroForge: Feedforward Text-to-Shape \\ Without 3D Supervision}

\author{%
  Kelly O. Marshall$^{\ast}$, Minh Pham$^{\ast}$, Ameya Joshi$^{\ast}$, Anushrut Jignasu$^{\dagger}$, Aditya Balu$^{\dagger}$, \\ \textbf{Adarsh Krishnamurthy}$^{\dagger}$\textbf{, Chinmay Hegde}$^{\ast}$ \\
  $^{\ast}$New York University \qquad $^{\dagger}$Iowa State University
}

\begin{document}

\maketitle

\begin{abstract}
Current state-of-the-art methods for text-to-shape generation either require supervised training using a labeled dataset of pre-defined 3D shapes, or perform expensive inference-time optimization of implicit neural representations. In this work, we present ZeroForge, an approach for zero-shot text-to-shape generation that avoids both pitfalls. To achieve open-vocabulary shape generation, we require careful architectural adaptation of existing feed-forward approaches, as well as a combination of data-free CLIP-loss and contrastive losses to avoid mode collapse. Using these techniques, we are able to considerably expand the generative ability of existing feed-forward text-to-shape models such as CLIP-Forge. We support our method via extensive qualitative and quantitative evaluations\footnote{Code available at: https://github.com/Km3888/ZeroForge}.
\end{abstract}

\section{Introduction}

\subsection{Motivation} 

High-quality 3D shape generation is an important primitive in numerous applications spanning video games, animation, scientific visualization, and the metaverse. Consequently, AI-based models for 3D shape generation have garnered significant recent attention. A large majority of such approaches pursue feedforward generative models (such as GANs), producing 3D shapes in various representations: pointclouds \citet{PointCloudGen,PointCloudGAN,PointFlow}, voxels \citet{3DGen}, and meshes \citep{PolyGen}\cite{AtlasNet}. A major drawback of the above class of approaches is that they require a labeled dataset of 3D shapes (such as ShapeNet~\citep{ShapeNet}), which typically only consist of a few categories. But for real-world applications, we seek models that can generate a wide variety---ideally, an open set---of shapes, succeeding with scarce (few-shot) or even no (zero-shot) labeled training data.

A solution to the data scarcity problem is by leveraging \emph{vision-language} pretraining. Models such as CLIP~\citep{CLIP} are pretrained on web-scale data to jointly learn correlated embeddings for both the visual features of an object and its textual description. Such models demonstrate spectacular zero-shot generalization capabilities, and have recently led to breakthroughs in image classification \citep{VirTex, DeViSE, Visual_Features}, retrieval \citep{CLIP2Vid, Fragment_Embeddings}, and image generation \citep{DALL-E, CLIPDraw, Text2Shape}. 

Starting from \citet{Parts2Words}, several recent efforts have adapted vision-language models (VLMs) for text-to-shape generation. Two important, zero-shot approaches along this direction are CLIP-Forge~\citep{CLIP-Forge} and Dream Fields~\citep{DreamFields}. CLIP-Forge is a feed-forward network trained on the 3D ShapeNet dataset to learn a text-to-shape mapping. It uses a normalizing flow model conditioned on CLIP text encodings coupled with a geometry decoder to produce diverse shapes represented by voxel grids. However, \emph{CLIP-Forge generalizes poorly} outside the categories in ShapeNet. This deficiency is addressed by Dream Fields, which trains a Neural Radiance Field (NeRF) for each object using only the supervision provided by an open-vocabulary natural language caption, plus a sparsity-based loss to encourage coherent geometries. However, NeRFs only implicitly represent shapes and are not generative feed-forward models; synthesizing even a single new concept with \emph{NeRFs require expensive inference-time optimization}, i.e., learning the weights of a  network from scratch.

\begin{figure*}[t!]
    \centering
    \includegraphics[width=0.99\linewidth]{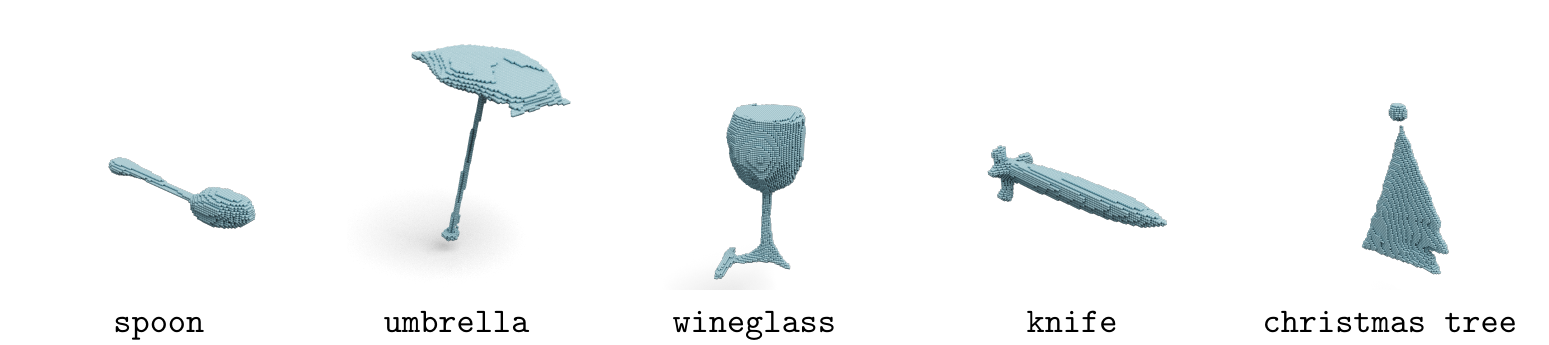}
    \caption{Example 3D shapes (above) with corresponding text inputs (below) from our proposed feedforward 3D generative text-to-shape model, ZeroForge. We are able to generate 3D voxel grids from arbitrary text queries without requiring 3D labeled shape data or NeRF-style optimization.}
    \label{fig:top_fig}
\end{figure*}


\subsection{Our Contributions} 

We present \textbf{ZeroForge}, a significant extension of the CLIP-Forge architecture for 3D shape generation that resolves the above two pitfalls, thereby achieves the best-of-all-worlds:
\begin{itemize}[left=0pt]
\item ZeroForge demonstrates considerable shape generalization abilities beyond ShapeNet.
\item ZeroForge does not require any 3D shape supervision, and instead trained using (frozen) CLIP embeddings of open-vocabulary, natural language captions.
\item ZeroForge incurs considerably lesser inference-time costs as compared with Dream Fields.
\end{itemize}

For the above reasons, we envision that ZeroForge can be used in several applications: quickly generating new families of paired shape-image datasets (beyond ShapeNet); 3D visualization of novel concepts described using natural language; probing geometric properties of specific 3D shapes using their voxel representations; and other related tasks in 3D vision. We present a host of 3D visualization results in support of ZeroForge. We also report the competitive performance of ZeroForge in terms of quantitative human evaluation metrics. See \secref{Sec:Results} for details.

\subsection{Techniques}

At a high level, the core ideas behind ZeroForge are conceptually very simple. Start with a pre-trained CLIP-Forge architecture (with a frozen CLIP text encoder); feed a novel text prompt through CLIP-Forge, render the output shape into 2D, and compute the embeddings of the resulting images (with a frozen CLIP image encoder); compare the similarity between the text and image embeddings; and use this as training signal to update CLIP-Forge decoder and flow model weights. 

However, care must be taken to make things work. The cosine similarity used by itself (without additional regularization) leads to mode collapse. Therefore, we train using batches of prompts, and augment the training objective with a contrastive loss term that encourages generator outputs to have higher similarity with their corresponding prompts than with all other text inputs. CLIP-Forge produces binarized voxel grids, which leads to non-differentiability. Instead, we use approximate binarization, coupled with a neural voxel renderer (NVR) to enable backpropagation of gradients from the similarity objective to the decoder weights. Finally, training on only new prompts (outside ShapeNet) leads to forgetting phenomena. We find that training convergence---as well as retention of \emph{original} ShapeNet generation capability---is improved by adding a zero-convolution adapter to the decoder, similar to the very recent work of~\citet{ControlNet}. See \secref{Sec:Methods} for details.

\section{Background}
\label{gen_inst}

\paragraph{Multi-modal Learning:}  Multi-modal Learning (MML) has been an important research area in recent decades. The objective of this approach is to enhance learning and predictive performance by integrating information from multiple sources. \citet{mdl_ngiam} focused on combining audio and visual data and demonstrated that when more than one modality is present during learning, deep networks can learn to leverage the correlative information, leading to improved performance in situations where one or more modalities may be missing. \citet{mdl_survay_baltrusaitis} focused on human communication and developed a method for understanding it as a combination of verbal (speech) and non-verbal (gesture, facial expression, etc.) signals. Their work underscored the importance of multi-modal learning in real-world applications such as human-computer interaction and affective computing.

Recently, vision-language model (VLM) learning has attracted a lot of attention from various research communities. Depending on the downstream task, different model architectures have been proposed, including the dual-encoder architecture \citep{CLIP, NoisySupervision}, the unified transformer architecture \citep{BLIP, image_foreign}, the fusion-encoder architecture \citep{LXMERT, align_before_fuse}, and the encoder-decoder architecture \citep{unifying_vl, SimVLM}   Additionally, various pre-training objectives have also been studied, which have progressively converged to a few time-tested ones: image-text contrastive learning \citep{CLIP, FILIP}, image-text matching \citep{align_before_fuse}, and (masked) language modeling \citep{Caption_Annotations}. VLMs have found considerable success in various tasks such as (zero-shot) classification \citep{VirTex, DeViSE, Visual_Features}, retrieval \citep{CLIP2Vid, Fragment_Embeddings}, and image generation \citep{DALL-E, CLIPDraw, Text2Shape}. 

\paragraph{CLIP-Forge:} A core component of our method is the CLIP-Forge model, which uses the ShapeNet dataset to learn a text-to-shape mapping \citep{CLIP-Forge,ShapeNet}. The CLIP-Forge pipeline is fully feed-forward and, once trained, requires no additional optimization to generate new shapes belonging to the classes in the dataset. This dataset can be represented as $\mathcal{S}= \left\{\left(  \textbf{V}_n, \textbf{I}_n, \textbf{O}_n  \right)\right\}_{n=1}^N$ where $V_n$ is a voxel grid of fixed resolution, $I_n$ is a rendered image of the shape, and $O_n$ is a set of space occupancies. We note that in addition to voxels, CLIP-Forge can also generate a point cloud, which we do not consider in our paper and leave as future work.

Due to the relatively small amount of 3D shape data, CLIP-Forge uses a multi-stage training process similar to \citep{VQVAE,HRSynthesis,TamingTransformers}. The first step is learning a latent distribution over all the shapes in the training set. This uses a voxel encoder $E$, which produces latent code $z_i=E(x_i)\in \mathbb{R}^h$. These latent codes are then passed to an Occupancy-Net~\citep{OccupancyNetworks} like decoder $D$, which predicts $\textbf{O}_n$, producing a reconstruction loss. In the second stage, a latent-flow model $\mathcal{F}$~\citep{NVP} learns the conditional distribution of latent codes $z_n$ based on the CLIP embedding of the rendered image $I_n$ . 

During inference, the image embedding is replaced by the CLIP embedding of a text query to give a sample from the distribution over latents conditioned on the text, which can be decoded to predict occupancy values at any points specified. To get a voxel-grid-shaped output, one can simply choose the grid vertices as the occupancy points. These occupancy predictions are then converted into binary voxel values using a constant threshold parameter $\gamma$ (typically 0.05).

\paragraph{NeRFs:} An emerging class of approaches implicitly represents 3D shapes in the form of neural radiance fields (NeRFs)~\cite{NeRF}. Original work compared rendered images to ground truth images, but recent work uses vision-language models to do text-to-shape modeling~\citep{DreamFusion,DreamFields,ClipNerf}. These can generate visually compelling outputs but require a new network for each shape. This necessitates training a network from scratch for each new text prompt. Such an ``inference-time'' computational cost makes these approaches unsuitable for generating arbitrary shapes on demand. 

\paragraph{Differentiable Rendering:} Using CLIP supervision to learn 3D shape generation requires that we backpropagate from RGB images into their corresponding shapes. To do this, we must define a rendering layer $\omega: \mathbb{R}^{N^3} \times \left[0,360 \right]^2 \rightarrow \mathbb{R}^{3 \times N^{2}}$ which maps a 3-dimensional voxel grid (along with a perspective angle) to its 2D rendering while also having a well-defined gradient. 

\citet{PlatoGAN} leverage image supervision to learn a 3D Generative Adversarial Network (GAN) by using a differentiable formulation of classical ray-tracing \citep{RayTracing} algorithms. However, these renderings struggle to capture object depth and lighting effects. Later works~\citep{NVR} have achieved better results by training a neural network model which is differentiable by design on large amounts of shape-rendering pairs to predict the images from 3D input.

\paragraph{ControlNet Finetuning:} With the widespread success of large foundation models, a recent area of research has focused on enabling the efficient adaptation of large models for specialized purposes \citep{VersatileTransfer,LORA}. A recent such method is ControlNet~\citep{ControlNet}, which addresses the specific problem of introducing conditional control to pre-trained diffusion models~\citep{Diffusion}. ControlNet learns a modified version of an original diffusion model by augmenting certain layers. Consider an arbitrary layer written as:
\begin{equation}
y=\mathcal{M}(x,\theta)
\end{equation}
with input $x$ and parameters $\theta$. To get an easily adaptable version of this layer, ControlNet makes a locked copy of the parameters $\theta_L$, which retains the capabilities of the original trained model as well as a trainable copy $\theta_C$. To allow conditioning on an additional input, this finetuned model accepts an additional parameter $c$. To combine these, they introduce a pair of $1 \times 1$ convolution kernels (that they call zero-convolution, or ZeroConv) with parameters $\theta_z1, \theta_z2$ initialized to zero and define a new controllable version of the output:
\begin{equation}
y_c = \mathcal{M}(x,\theta_L) + \mathcal{Z}(\mathcal{M}(x+\mathcal{Z}(c;\theta_{z1},\theta_C),\theta_{z2})
\label{eq:zero_conv}
\end{equation}
Because the convolution layers are initialized with zero entries, we have that $y_c=y$ for all inputs at the beginning of training. However, introducing additional trainable parameters allows the network to rapidly adapt to the new finetuning task. This is achieved while maintaining performance on the original pre-training task, as at each layer, the convolution operator can filter out the output from the finetuned parameters.

\section{Methods}
\label{Sec:Methods}

\begin{figure*}[!b]
    \centering
    \includegraphics[width=0.99\linewidth]{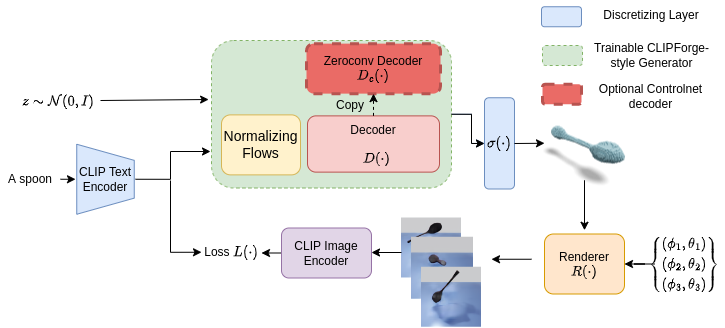}
    \caption{\textbf{ZeroForge Block Diagram.} Illustration of our novel method for learning text-to-shape models. Outputs from a pre-trained CLIP-Forge model are rendered into images and encoded into CLIP space. By computing their similarity with the input text queries we obtain a tractable loss function for arbitrary text queries}
    \label{fig:zf_block}
\end{figure*}

Our method ZeroForge (see \figref{fig:zf_block}) can be viewed as an adaptation of CLIP-Forge to generate an arbitrary distribution over text queries $p(t)$. We begin by using pre-trained CLIP-Forge architectures that have been trained on the ShapeNet dataset; these are well-suited text-to-shape generation for the classes within this dataset. Using this initialization gives a sensible starting point from which the networks can learn novel 3D concepts of interest to us.

To enable generalization to concepts outside ShapeNet, we first leverage a similarity loss based on maximizing an inner product between image and (open vocabulary) text embeddings provided by CLIP~\citep{CLIP}. We also define an additional contrastive loss term to discourage mode collapse. To minimize these loss terms, we formulate a differentiable layer for getting binary voxel occupancies from the model's real-valued outputs. Finally, we describe how we can apply the fine-tuning methods used in ControlNet \cite{ControlNet} to rapidly learn new shapes while preserving performance on shapes from the ShapeNet dataset.

\subsection{Similarity Loss}

For performing our adaptation, we can set aside CLIP-Forge's encoder $E$ and define our generator $G = D \circ \mathcal{F}$ as the composition of the latent flow network and voxel decoder (In contrast, recall that the CLIP-Forge model is $E \circ D$). We define a new training objective for this generator which measures how well the renderings of its voxel grid outputs match the text descriptions given as input. To do this, we sample angles randomly from the distribution $q(\theta,\phi)$ over the unit sphere using \citet{SphereSampling} and differentiably render the shape from these angles using the renderer $\omega$~\citep{NVR}. These 2D images can then be embedded into CLIP space using CLIP's image encoder head and compared to the encoded text input. This allows us to define the similarity loss $\mathcal{L}_{sim}$ which encourages voxel outputs matching the text queries drawn from $p(t)$:
\begin{equation}
\mathcal{L}_{sim}(G) = -\mathbb{E}_{t \sim p,(\theta,\phi) \sim q} \Big[ h(\omega(G(g(t)), \theta,\phi)) \cdot g(t) \Big] \, .
\end{equation}

\subsection{Contrastive Loss}
While this simple similarity loss can achieve 3D shapes which have high similarity in CLIP embedding space, we observe that using it by itself can result in mode collapse. This occurs when the input distribution contains text queries that have high similarity with one another in CLIP space. In this scenario, the generator will converge to the degenerate solution of producing whichever shape is simplest, regardless of the input query. For instance, the queries \texttt{``spoon''} and \texttt{``fork''} have very high similarity, despite these shapes having very different appearances. As a result, if the generator outputs spoons for the input \texttt{``fork''}, it will still partially satisfy the similarity loss.

To remedy this, we augment $\mathcal{L}_{sim}$ a contrastive loss term based on InfoNCE~\citep{InfoNCE} to our training objective. This encourages each of the generator's outputs to have higher similarity with its corresponding text input than with all the other inputs. We define this loss for a batch of size $B$ with $\mathcal{I},\mathcal{T} \in \mathbb{R}^{B \times H}$ representing the image and text embeddings, respectively:

\begin{equation}
\mathcal{L}_{contrast}(\mathcal{I},\mathcal{T},\tau) = \frac{1}{M}\sum_{i=1}^{B} \log \left( \frac{\exp( \tau \mathcal{I}_i \, \mathcal{I}_i)}{\sum_{j=1}^{B} \exp(\tau \mathcal{I}_j \, \mathcal{T}_i)} \right) ,/ .
\label{eq:contrast}
\end{equation}

This loss prevents mode collapse by penalizing similarity between text-image pairs that do not correspond to one another. The temperature hyperparameter $\tau$ controls how steeply the penalization is distributed amongst the different non-matching text queries. When $\tau$ is high, only the text queries with the highest similarity are punished, and when it is zero, all queries are punished equally regardless of similarity to the rendered image. We add this contrastive loss to our original training objective as a regularizer multiplied by the hyperparameter $\lambda_C$. We tune this hyperparameter to avoid mode collapse while allowing the generator to accurately generate shapes that are similar to one another.

\subsection{Differentiable Binarization}

Unlike CLIP-Forge, we do not fully binarize the output voxel grid based on whether the voxel values are above a fixed threshold. Instead, we use approximate Binarization using a sigmoidal transfer function $B$, producing a grid representation that is differentiable, following \citet{Binarization}. During training, we use the same threshold value $\gamma$ that was used for the CLIP-Forge initialization and get a near-binary voxel output.

\begin{equation}
B(x, \beta, \gamma) = \frac{1}{1 + \exp(\beta (x-\gamma))}
\label{eq:binarization}
\end{equation}

The temperature parameter, $\beta$, controls the tradeoff between a tight approximation and smooth gradients, with $B$ recovering the regular step function in the limit as $\beta \rightarrow \infty$. This ``soft thresholding'' allows us to backpropagate gradients through the binarization operation at train time while ensuring that our generator will produce sensible outputs when we use the regular thresholding operation during inference.

\subsection{ZeroConv}

We implement a variant of the zero-convolution layer from \citet{ControlNet} to improve training convergence and improve the tradeoff between the performance on the fine-tuning queries and those in the original ShapeNet dataset. Because we are not concerned with adding conditional control to the model, we can simplify \Eqnref{eq:zero_conv} to:
\begin{equation}
y_c = F(x,\theta_L) + \mathcal{Z}(F(\theta_{z})).
\end{equation}

We replace each residual block in the pre-trained decoder with an augmented version based on this formulation, with a 3D convolutional layer to connect the frozen CLIP-Forge weights and the fine-tuned weights learned using the loss we define. This additional set of weights can be referred to as $D_c$, as illustrated in \figref{fig:zf_block}.

\section{Results}
\label{Sec:Results}

\begin{figure}[b!]
\centering
\includegraphics[width=0.99\linewidth]{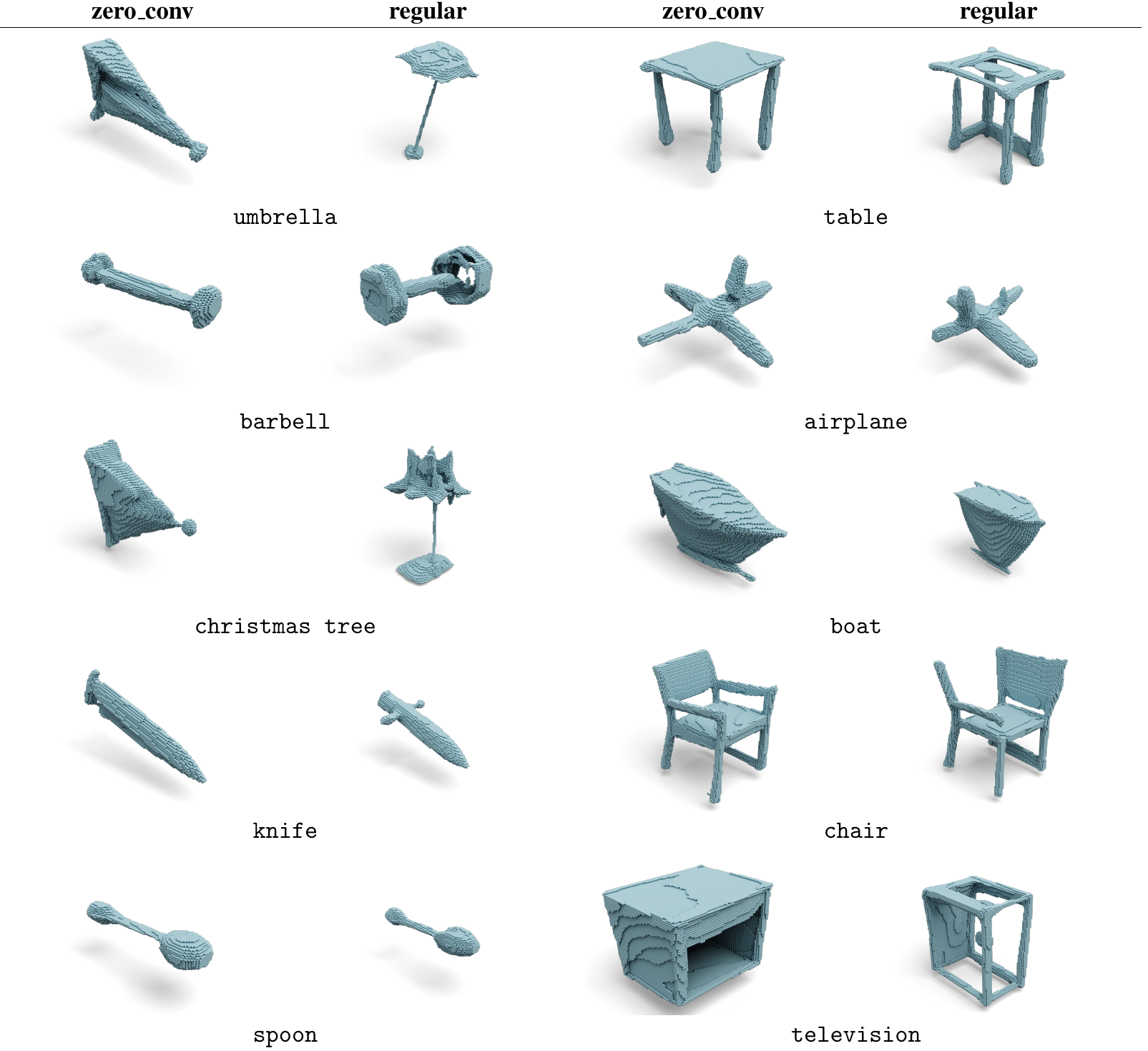}
\caption{Shapes generated from ZeroForge after finetuning on 5 queries which were not seen in pre-training data (umbrella, barbell, Christmas tree, knife, spoon) and 5 which were (table, airplane, boat, chair, television). Here we use $\lambda_C=0.01$}

\label{tab:5_with_original}
\end{figure}

\begin{figure*}[!t]
    \centering
    \includegraphics[width=0.99\linewidth,trim={0.0in 1.5in 0.0in 1.5in},clip]{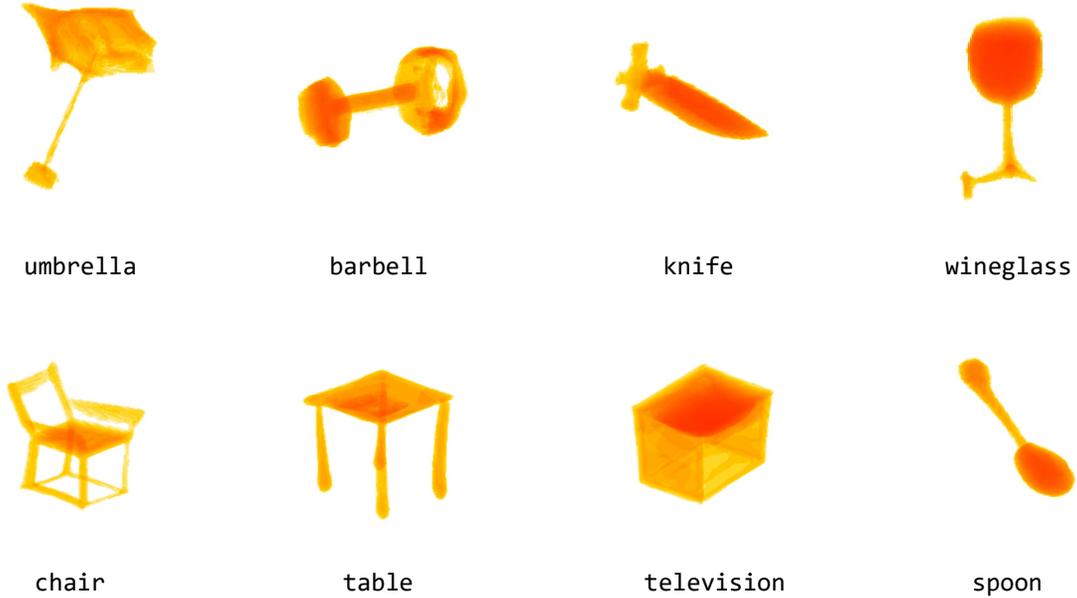}
    \caption{Volume renderings for various shapes generated by ZeroForge (umbrella, barbell, knife, wineglass, chair, table, television, and spoon).}
    \label{fig:best_vol_renders}
\end{figure*}

\begin{figure}[b!]
\includegraphics[width=0.99\linewidth]{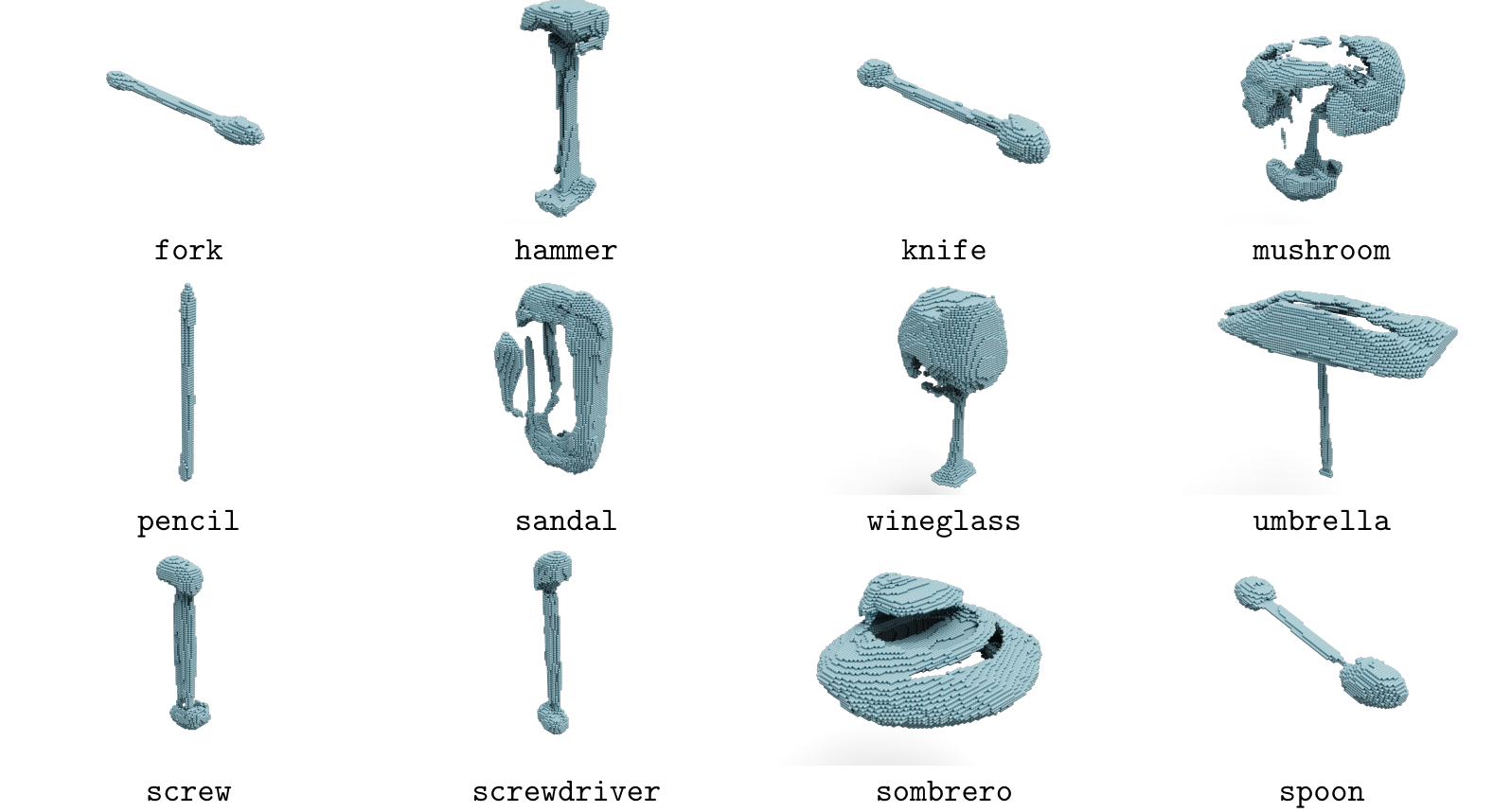}
\caption{Samples from a ZeroForge model trained on 12 challenging and complex shape classes not seen in the original data distribution.}

\label{tab:6_with_original}
\end{figure}

\subsection{Generating Novel Shapes}

To test our method, we perform qualitative evaluations on several distributions over text queries. First, we gradually expand the support of CLIP-Forge's distribution from the classes in ShapeNet by training ZeroForge on $10$ text queries. Five (Table, Airplane, Boat, Chair, Television) are present in the ShapeNet dataset, and the other five (Umbrella, Barbell, Christmas tree, Knife, Spoon) are novel. We set $\lambda_c=0.01, \tau=50$ and report results from both with and without the ZeroConv modification of the decoder network. We adopt the same voxel rendering technique as CLIP-Forge for visualization purposes.

The results of this experiment are shown in \figref{tab:5_with_original}. Both choices of decoder architecture allow ZeroForge to learn the task distribution with minimal mode collapse, capturing the majority of shapes. However, the addition of ZeroConv shows a clear improvement in shapes within the original ShapeNet distribution. Because the original CLIP-Forge parameters are frozen, and the convolution kernels can easily learn to filter out the results of the fine-tuned parameters at each layer, these shapes are largely identical to those that would have been produced by the CLIP-Forge model. For visual clarity, we also show the volume renderings of various shapes generated by ZeroForge in \figref{fig:best_vol_renders}. Volume renderings were generated using GPU-accelerated ray-marching of the voxel grid~\citep{fernando2004gpu}.

A key observation from our experiments is the importance of initializing ZeroForge with CLIP-Forge weights. Without this inductive bias, we find that gradient descent cannot find a sensible solution to the data-free shape generation process. Shapenet-based training of CLIP-Forge gives our network a solid foundation for representing basic primitive shapes, which it can learn to reform into new shapes that satisfy the CLIP-based loss. We demonstrate this with additional figures in the appendix, showing ZeroForge's attempt to learn the same ten shapes shown in \figref{tab:5_with_original} using the same architecture as before but with randomly initialized weights.

\begin{figure}[t!]
\includegraphics[width=0.99\linewidth]{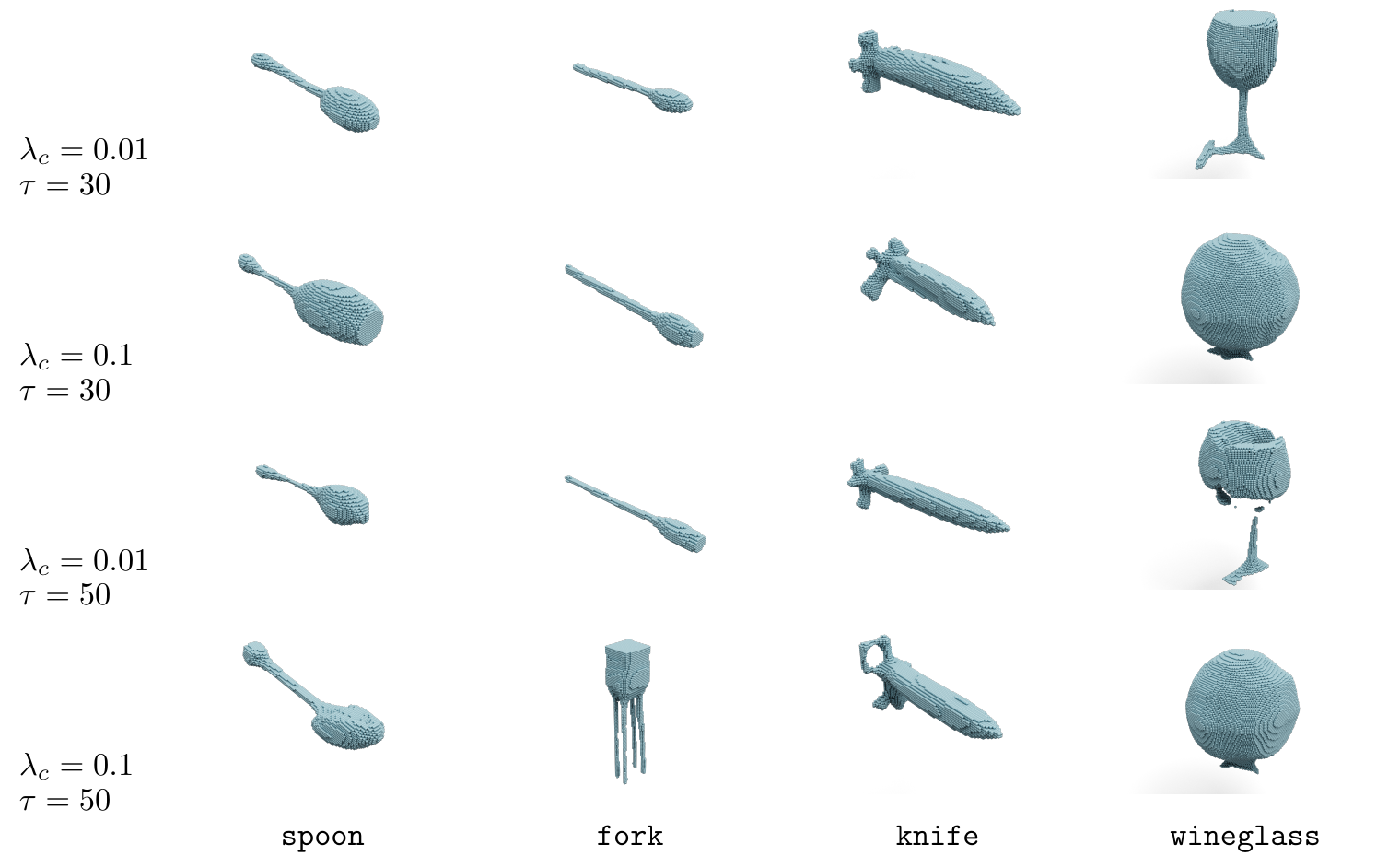}
\caption{Ablation study on shapes generated from ZeroForge using different values of $\lambda_c$ and $\tau$.}
\label{tab:contrast_ablation}
\end{figure}

To demonstrate the problem of mode collapse and how it can be addressed by the contrastive loss term defined in \Eqnref{eq:contrast}, we also provide an example of a set of text queries that are prone to collapse, seen in \Eqnref{tab:contrast_ablation}. These four queries \{\texttt{spoon}, \texttt{fork}, \texttt{knife}, and \texttt{wineglass}\} are all tableware and, as a result, have a significant commonality in CLIP's text embedding space. This makes it particularly difficult to avoid collapsing to a single-shape output in this case. As a result, we can observe the role of the $\lambda_C$ and $\tau$ hyperparameters. Even setting $\lambda_c=0.01$, which we find to be sufficient for other text queries, there is little difference between the outputs for \texttt{fork} and \texttt{spoon}. Getting the model to produce the fork tines requires not only increasing the value of $\lambda_c$, but also the $\tau$ to properly penalize the errant similarity between the renderings of the fork and spoon outputs.

To test ZeroForge's limits regarding the size of the query distribution, we train ZeroForge on a set of 12 challenging shape descriptions, all of which are outside the original CLIP-Forge dataset. These queries are: \{ \texttt{fork}, \texttt{hammer}, \texttt{knife}, \texttt{mushroom}, \texttt{pencil}, \texttt{sandal}, \texttt{screw}, \texttt{screwdriver}, \texttt{sombrero}, \texttt{spoon}, \texttt{umbrella}, \texttt{wineglass} \}.

To measure how well the ZeroForge-generated shapes appear to human beings, we conducted a human perceptual evaluation via an online survey. We mirrored the protocol outlined in the paper describing CLIP-Forge~\cite{CLIP-Forge}. Participants were shown a pair of shapes generated by ZeroForge and asked to determine which of the two objects matched the text query used to generate one of them. For this experiment, we used the objects displayed in \figref{tab:5_with_original} and the model trained using a ZeroConv decoder. A total of 65 participants were queried, with each one judging 17 pairs of queries.

\begin{table}[t!]
\setlength\extrarowheight{3pt}
\centering
    \begin{tabular}{ | l |l |l |}
    \hline
     In-distribution & Out-of-Distribution & Overall \\
    \hline
    95.0$\pm11.9$ \%          & 95.7$\pm 13.3$\%              & $95.4\% \pm 5.0\%$  \\
    \hline
    \end{tabular}
    \medskip
    \caption{Accuracy of ZeroForge in a human evaluation survey.}
    \label{tab:survey}
\end{table}

The average accuracy was $95.4\%$ (\tabref{tab:survey}) across all queries, with over half of surveyed participants identifying the correct answer for all 17 questions. The standard deviation of the accuracy rate across participants was $5$ percentage points. Subdividing this accuracy between shapes that were in the ShapeNet dataset and those which were not, we get that the former grouping yielded an average of $95.0\%$ correct, and the latter gave an average of $95.7\%$ correct, well below the threshold of statistical significance.

Please see the appendix for several more experimental results and ablation studies.

\section{Conclusions}

In this paper, we presented ZeroForge, a novel feedforward text-to-shape model capable of generating open-vocabulary shapes (in terms of their voxel grid representation) without 3D labeled supervision. We validated ZeroForge using several qualitative and quantitative evaluations. 

Several avenues for future work remain. We have focused on voxel grid representations for generated shapes, but extensions to point clouds and mesh-based representations are a natural next step. A better understanding of the impact of various architectural choices: pre-training weights, prompt context length, and the capacity of flow model and decoder, would pave the way for better zero-shot 3D generation. Finally, a hybrid combination with NeRF-style methods would enable high spatial resolution, as well as endow generated shapes with improved texture properties.

\subsection{Limitations}
\label{Sec:Limitations}

Despite the several encouraging results, we find several key areas in which future work could improve upon our results. 

One such area is in improving the tradeoff introduced by the contrast coefficient. While higher values of this hyperparameter are necessary for preventing mode collapse and allowing the generation of certain shapes, it can also result in dissimilarity amongst shapes which ought to be similar. This can be seen in \figref{tab:contrast_ablation}, where to avoid generating spoons in place of forks, the contrastive loss degrades the wineglass shape to avoid it resembling a spoon.

Our results also show that the inductive bias provided by using pre-trained CLIP-Forge weights as initialization for ZeroForge can also affect the performance of the final trained ZeroForge model. For instance, even when ZeroForge avoids mode collapse for the \texttt{fork} query, its output still somewhat resembles a chair, as this is part of the ShapeNet dataset that CLIP-Forge is trained on. This can be seen in the training process as early on, the outputs for queries not seen in the training data look almost identical to ShapeNet objects and then \emph{continuously deform} into the correct shapes. This raises interesting questions of whether ZeroForge can generate shapes with completely new (and intricate) topological properties, and we defer a more thorough characterization of this question to future work. 

In several of the generated shapes, there are visible holes and areas of the shape not filled in. This is likely attributable to the differentiable binarization layer (\Eqnref{eq:binarization}), which assigns non-zero occupancy values to the generated voxels just below the thresholding value $\gamma$. If there are several such voxels in the same area, they can form a low opaqueness region which is rendered as part of the shape. This can be addressed by increasing the binarization temperature $\beta$, but doing so also adversely affects our optimization procedure as it causes the loss gradients to become less smooth. Another potential method for improving this would be to use a smoothing method such as bilateral filtering \cite{BilateralFiltering} instead of a fixed threshold value.

Finally, like CLIP-Forge, we do not modify the CLIP text encoder and therefore are limited by CLIP's context length (currently, 77 tokens). But we envision that text encoders from future vision-language models with longer context lengths can be integrated within ZeroForge to resolve this limitation.

\subsection{Broader Impacts}

By building and validating ZeroForge, we have demonstrated rapid (feedforward) generative capability of complex 3D shapes using very weak (natural language) supervision. This opens the door to several new computational tools: interactive 3D visualization; understanding geometric properties by probing their voxel representations; and novel dataset generation and curation. These tools open the door for several fruitful applications, including improved design and manufacturing of novel industrial tools, as well as visualizing new material geometries. 

However, more sophisticated versions of ZeroForge could also be used to realistically hallucinate 3D objects, and, combined with augmented reality tools, could be used as a misinformation tool when used inappropriately. On balance, we feel that techniques such as ZeroForge pave the way for a better overall understanding of the scope of modern 3D shape generation tools, which is a necessary step for their safe deployment. 

\section*{Acknowledgements}

This work was supported in part by NSF CAREER grants CCF-2005804 and OAC-1750865, NSF grant LEAP-HI-2053760, and NSF/USDA-NIFA grant 2021-67021-35329. KOM was also supported by a US Department of Education GAANN Fellowship.

\clearpage

\appendix

\section{Appendix}


\subsection{Implementation Details}

Architecture:

Our model architecture built off of the default implementation provided by CLIP-Forge\citep{CLIP-Forge}. The RealNVP \citep{NVP} latent flow model consisted of 5 coupling layer blocks, each with translation and scaling followed by batch norm and using the inverse of the masking scheme from the previous block. The blocks themselves each contained 2-layer MLPs with a hidden size of $1024$ followed by an additional linear layer. The decoder model uses 5 3-dimensional ResNet decoder blocks. The shape embedding outputted by the flow model and given to the decoder has $128$ dimensions.

For training ZeroForge, we use an Adam optimizer with a learning rate of $10^{-5}$ and $\beta_1=0.9,\beta_2=0.999$. We set the batch size to be three times the number of queries to try to obtain accurate gradients. All ZeroForge visualizations come from models trained for 15k training iterations unless otherwise noted.

Hparams:
Unless otherwise specified we use the following hyperparameters in each ZeroForge experiment: $\beta=200, \tau = 50,\gamma=0.05$. We generate our voxels with a resolution of 128 in each dimension and resize our rendered images to $224 \times 224$ pixels in each dimension to make them inputtable to CLIP. All models for experiments were trained on 4 NVIDIA A100 GPUs.

\subsection{Additional Experiments}

\begin{figure}[b!]
\includegraphics[width=0.99\linewidth]{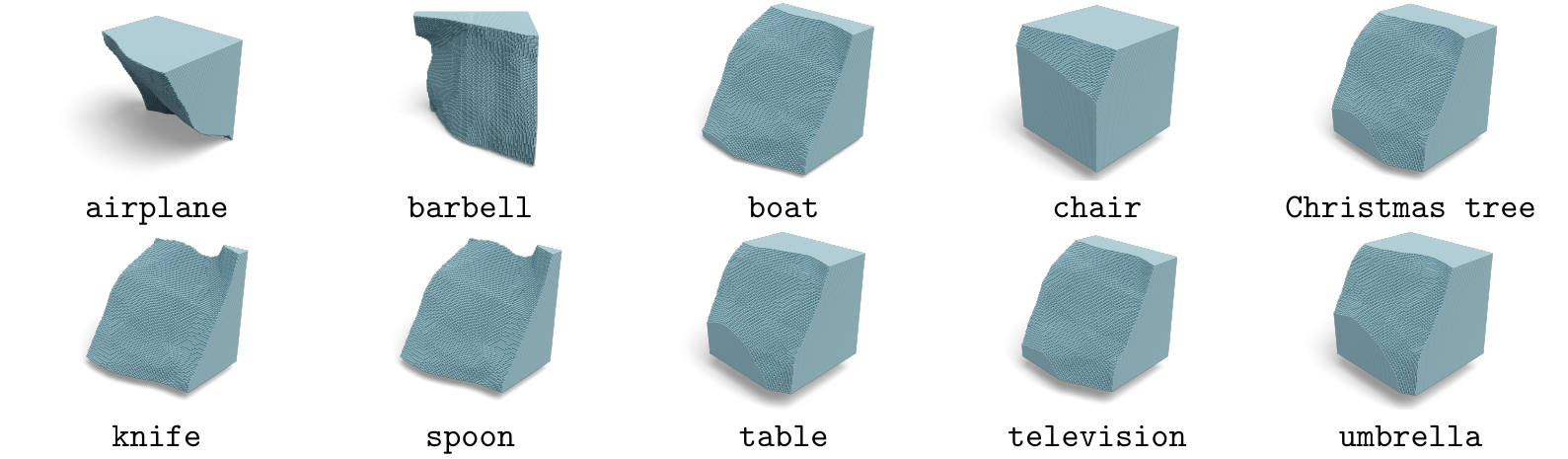}
\caption{Samples from a ZeroForge model trained without using CLIP-Forge weights as an initialization. Notice that these outputs lack definition as compared with typical ZeroForge outputs}
\label{tab:without_CLIP_Forge}
\end{figure}

The first set of additional experiments we present is a test of the importance of using CLIP-Forge weights to initialize our ZeroForge method. We mimic the training procedure and architecture used in training ZeroForge to produce the images in \figref{tab:5_with_original}, but instead of initializing the latent flow and decoder models using CLIP-Forge weights we instead initialize them randomly. We show in \figref{tab:without_CLIP_Forge} that without the inductive bias provided by CLIP-Forge's initialization, the networks are unable to generate sensible shapes using only the similarity-based CLIP loss. Instead, the model converges to a degenerate solution in which it outputs blockish shapes that lack the recognizable features of typical ZeroForge outputs. 

\begin{figure}[b!]
\includegraphics[width=0.99\linewidth]{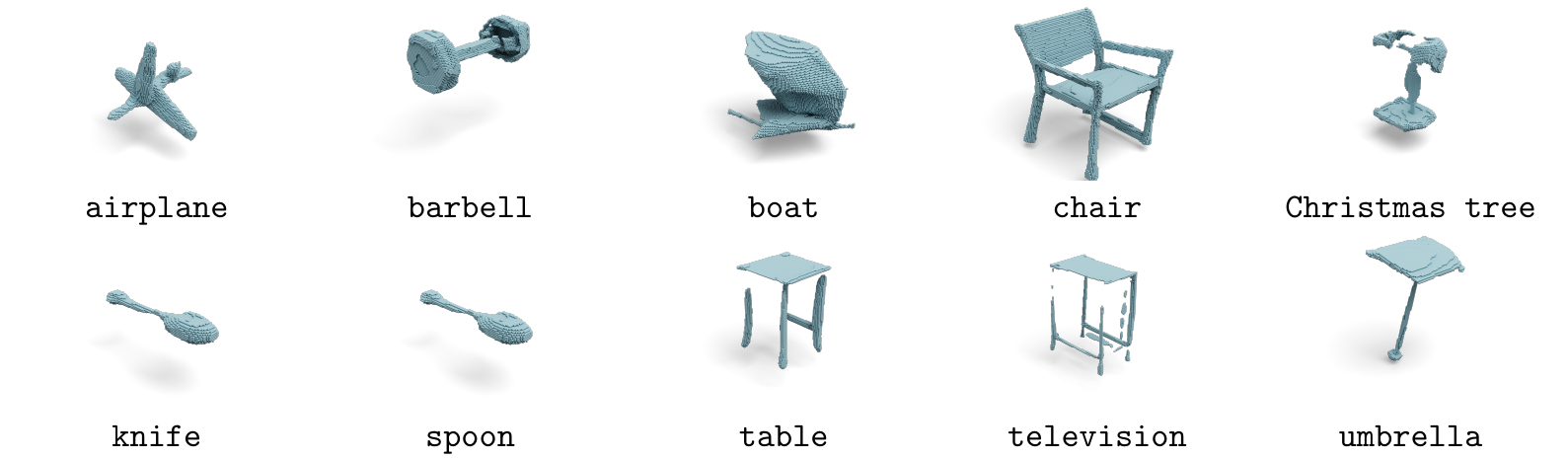}
\caption{Samples from a ZeroForge model trained with $\beta=100$}
\label{tab:beta=100}
\end{figure}

\begin{figure}[b!]
\includegraphics[width=0.99\linewidth]{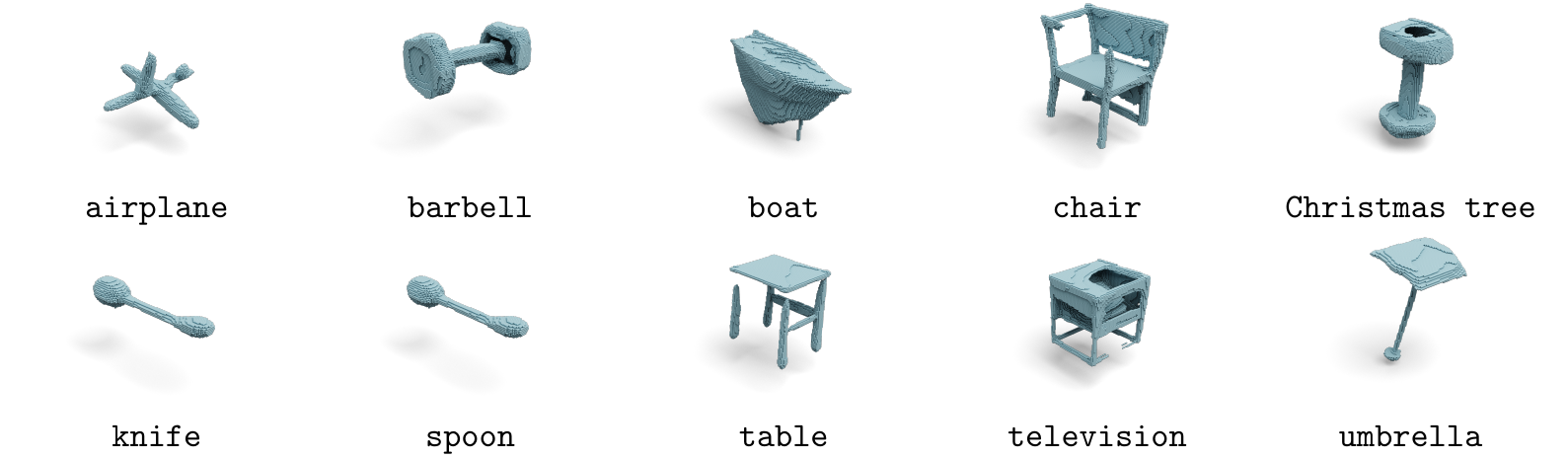}
\caption{Samples from a ZeroForge model trained with $\beta=300$}

\label{tab:beta=300}
\end{figure}

As an additional ablation, we examine the effect of the $\beta$ hyperparameter on ZeroForge's performance, showing the outputs of models trained using $\beta=100$ \figref{tab:beta=100} and $\beta=300$ in \figref{tab:beta=300} (our default setting for this hyperparameter was 200). We see that ZeroForge is relatively robust to this choice of hyperparameter, with distinguishable outputs still being generated at lower and higher values. We do however observe some degradation in quality at $\beta=100$ compared to the default setting and in training the model with $\beta=300$ there is a slower rate of convergence.

\begin{figure}[b!]
\includegraphics[width=0.99\linewidth]{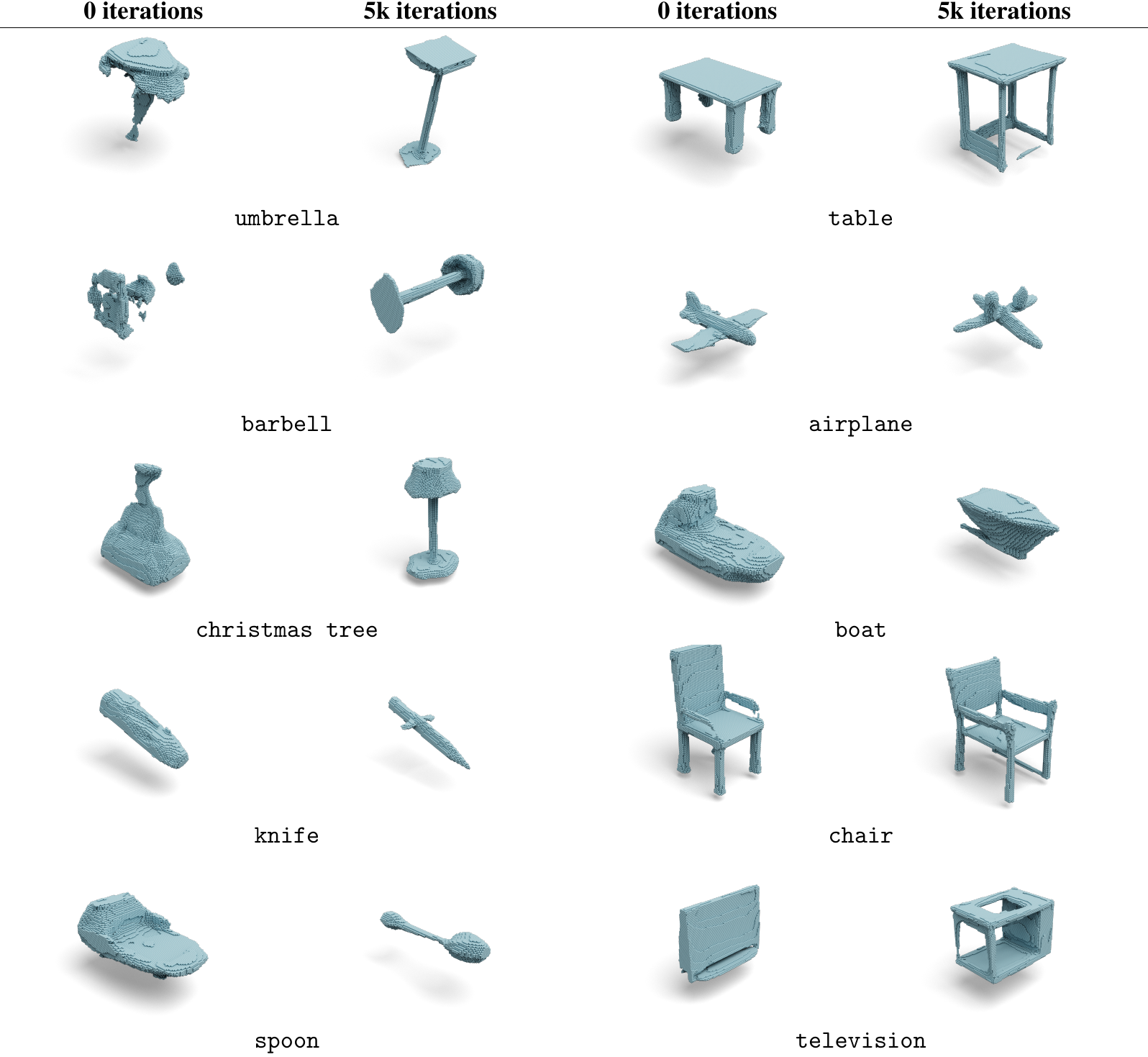}
\caption{Shapes generated by untrained and partially trained ZeroForge models. Note how several shapes outside CLIP-Forge's are initially represented by shapes inside the distribution which resemble them.}
\label{tab:partiall_trained}
\end{figure}

Finally, we demonstrate the progression of ZeroForge's outputs during training by including shapes generated by a partially trained model in \figref{tab:partiall_trained}. We select the model used to generate \figref{tab:5_with_original} after 0 iterations and 5000 iterations of ZeroForge training (note that the first is just the weights of the CLIP-Forge initialization). In this case we select the variant without ZeroConv so that that the progressive degradation on ShapeNet shapes can also be visualized. It can be seen here how ZeroForge learns to represent novel shapes by deforming and rearranging shapes that were in CLIP-Forge's training distribution. For instance, "barbell" and "Christmas tree" both start off resembling lamps as that is a shape present in ShapeNet.

\subsection{Visualizations}

For visual clarity purposes, we generate volume renderings for each of the shapes shown in \figref{fig:top_fig}, \figref{tab:5_with_original}, \figref{tab:6_with_original}, and  \figref{tab:contrast_ablation}.

\begin{figure*}[!t]
    \centering
    \includegraphics[width=0.99\linewidth,trim={0.0in 4.25in 0.0in 1.0in},clip]{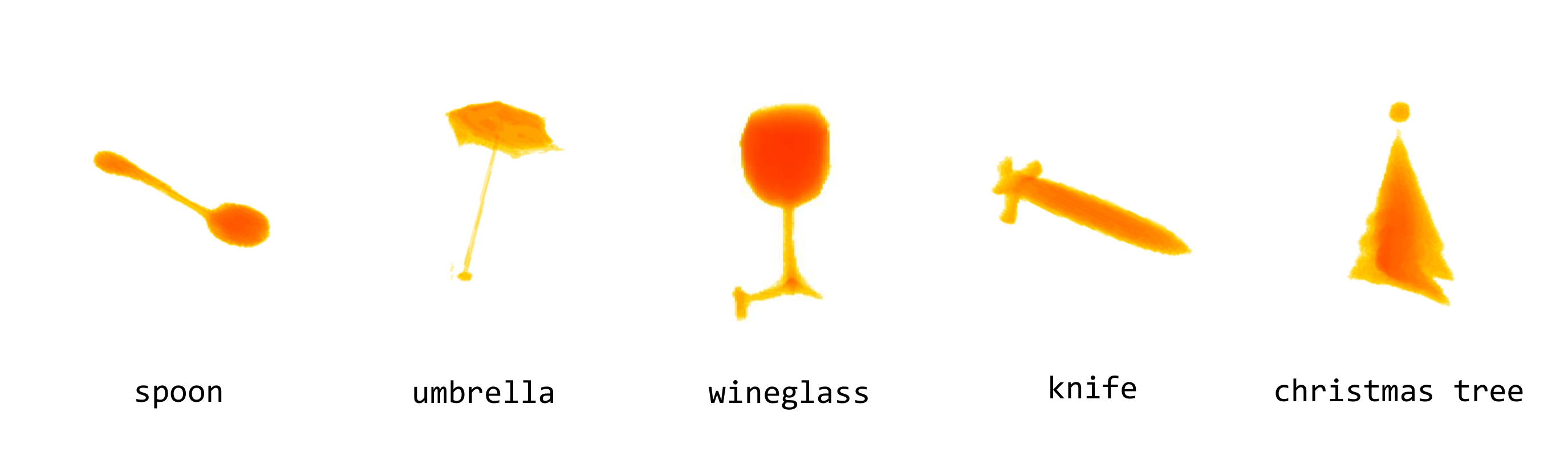}
    \caption{Volume renderings for shapes in \figref{fig:top_fig}.}
    \label{fig:fig1_vol_render}
\end{figure*}

\begin{figure*}[!t]
    \centering
    \includegraphics[width=0.99\linewidth,trim={0.0in 1.5in 0.0in 1.5in},clip]{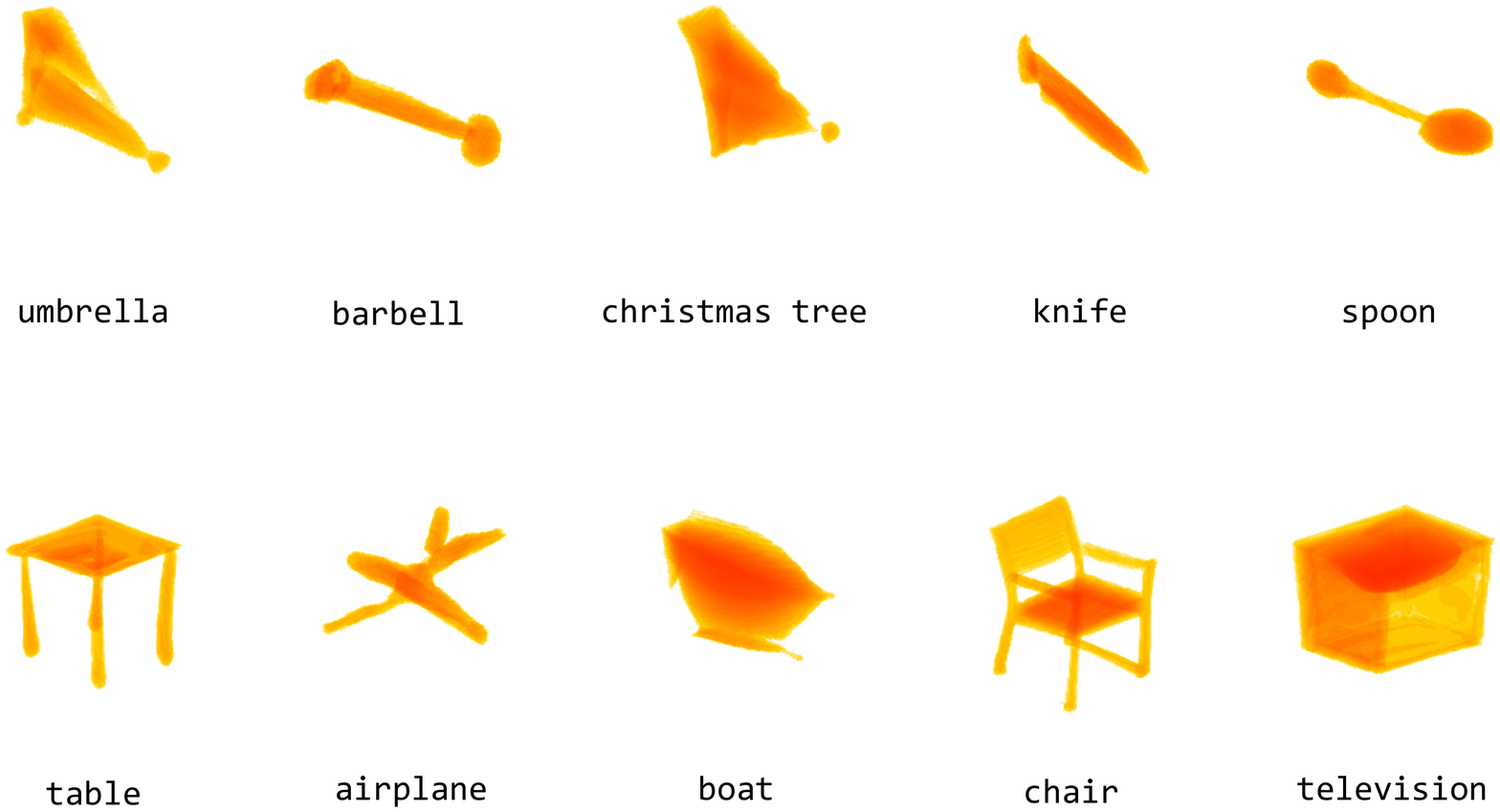}
    \caption{Volume renderings for \textbf{zero\_conv} config from \figref{tab:5_with_original}.}
    \label{fig:fig3_zc_vol_render}
\end{figure*}

\begin{figure*}[!t]
    \centering
    \includegraphics[width=0.99\linewidth,trim={0.0in 1.5in 0.0in 1.5in},clip]{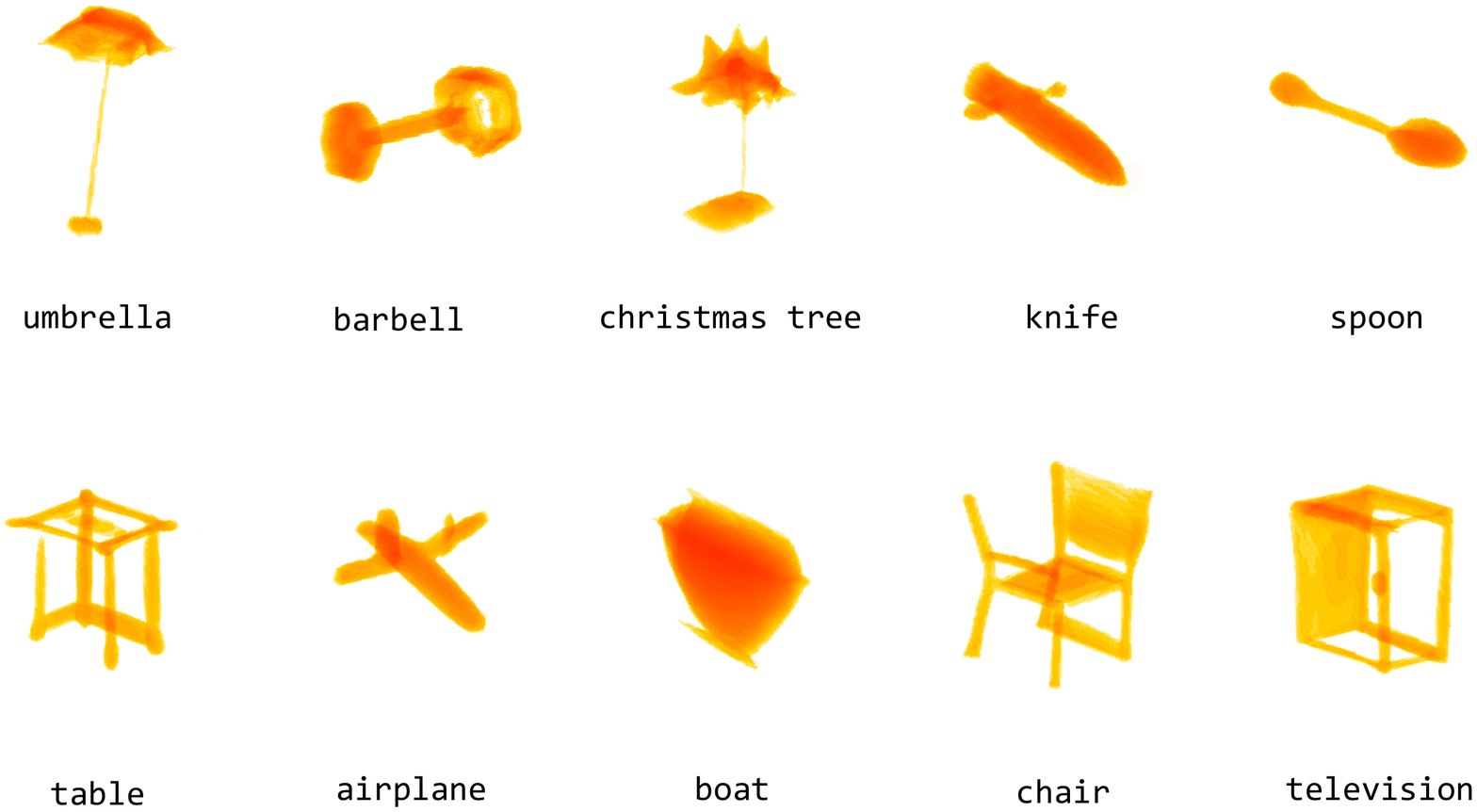}
    \caption{Volume renderings for \textbf{regular} config from \figref{tab:5_with_original}.}
    \label{fig:fig3_og_vol_render}
\end{figure*}

\begin{figure}[b!]
\centering
\begin{tabular}{c}
    \includegraphics[width=0.99\linewidth,keepaspectratio,clip,trim={0.0in 1.0in 0.0in 1.0in}]{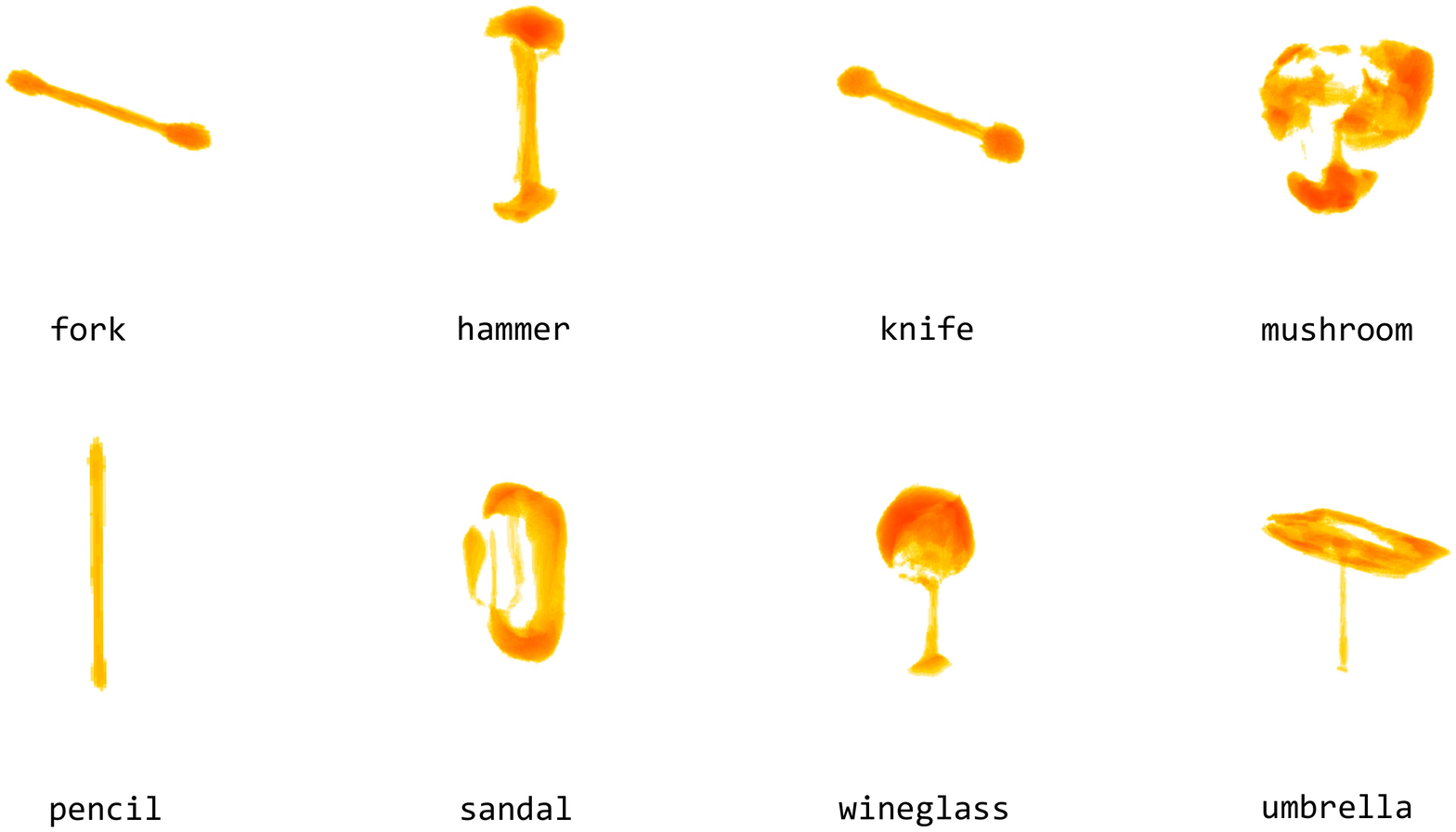} \\
    \includegraphics[width=0.99\linewidth,keepaspectratio,clip,trim={0.0in 4.5in 0.0in 1.5in}]{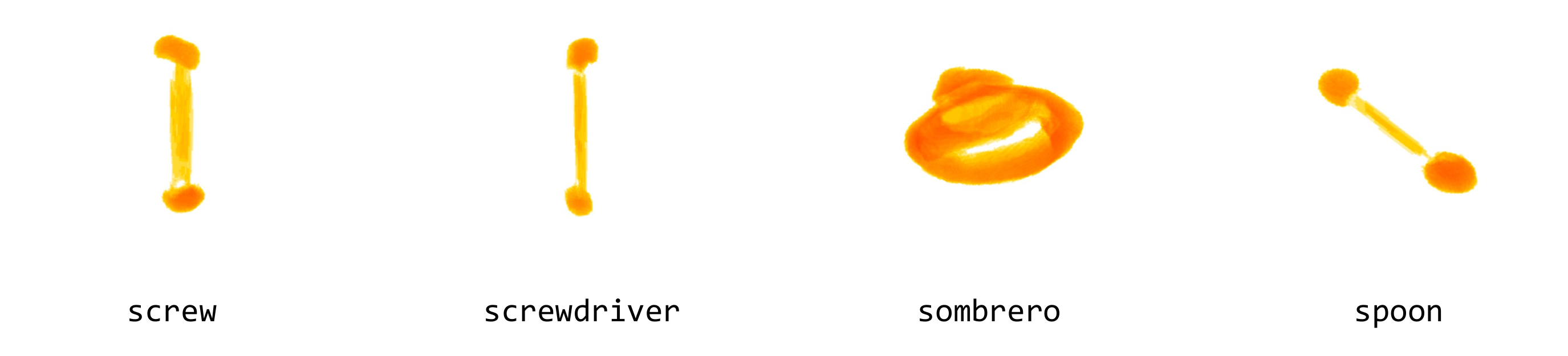} 
\end{tabular}
\caption{Volume renderings for shapes shown in \figref{tab:6_with_original}.}
\label{tab:fig5_vol_render}
\end{figure}

\begin{figure}[b!]
\centering
\begin{tabular}{c}
    \includegraphics[width=0.99\linewidth,keepaspectratio,clip,trim={0.0in 1.5in 0.0in 1.0in}]{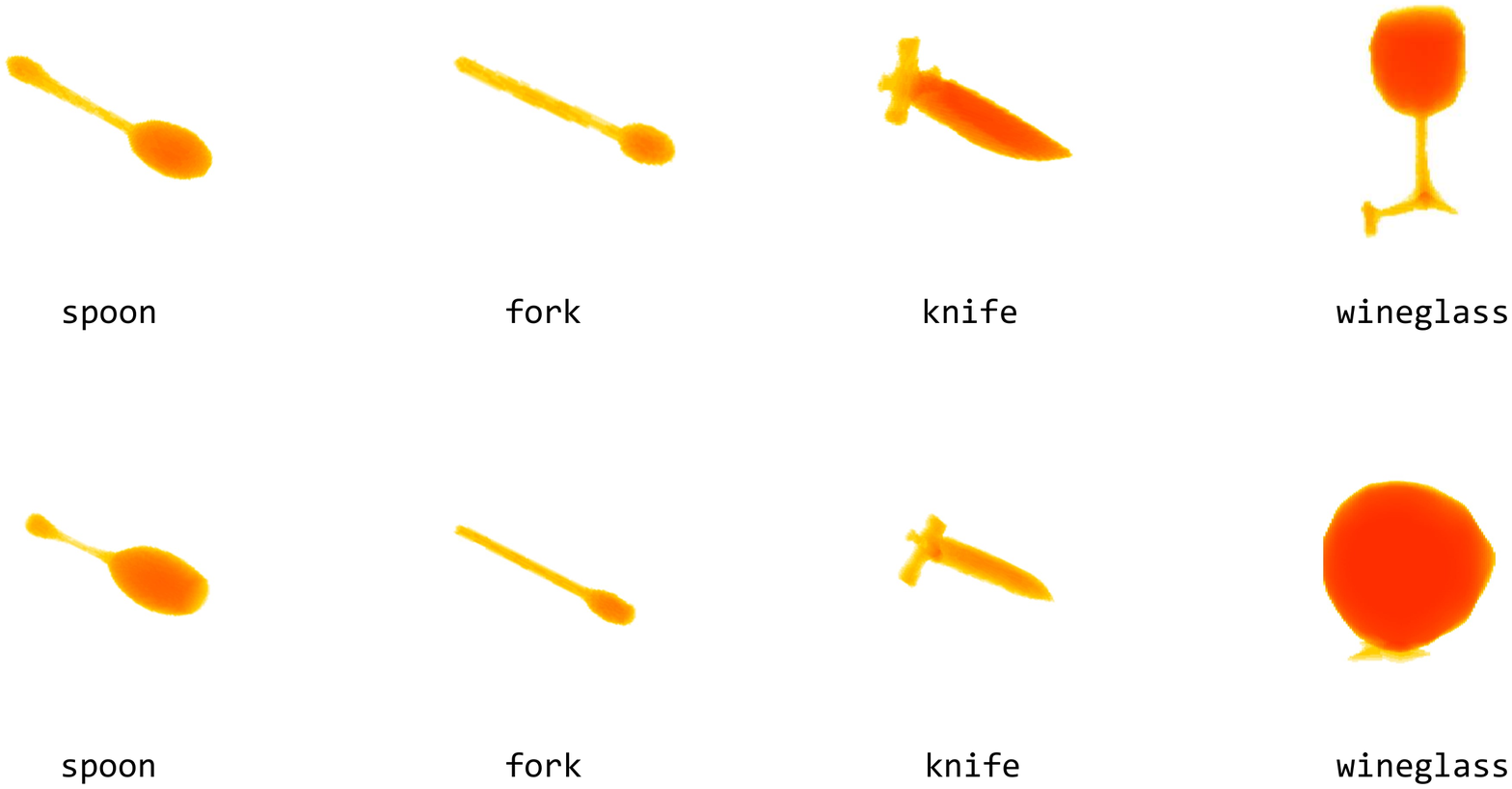} \\
    \includegraphics[width=0.99\linewidth,keepaspectratio,clip,trim={0.0in 1.5in 0.0in 1.5in}]{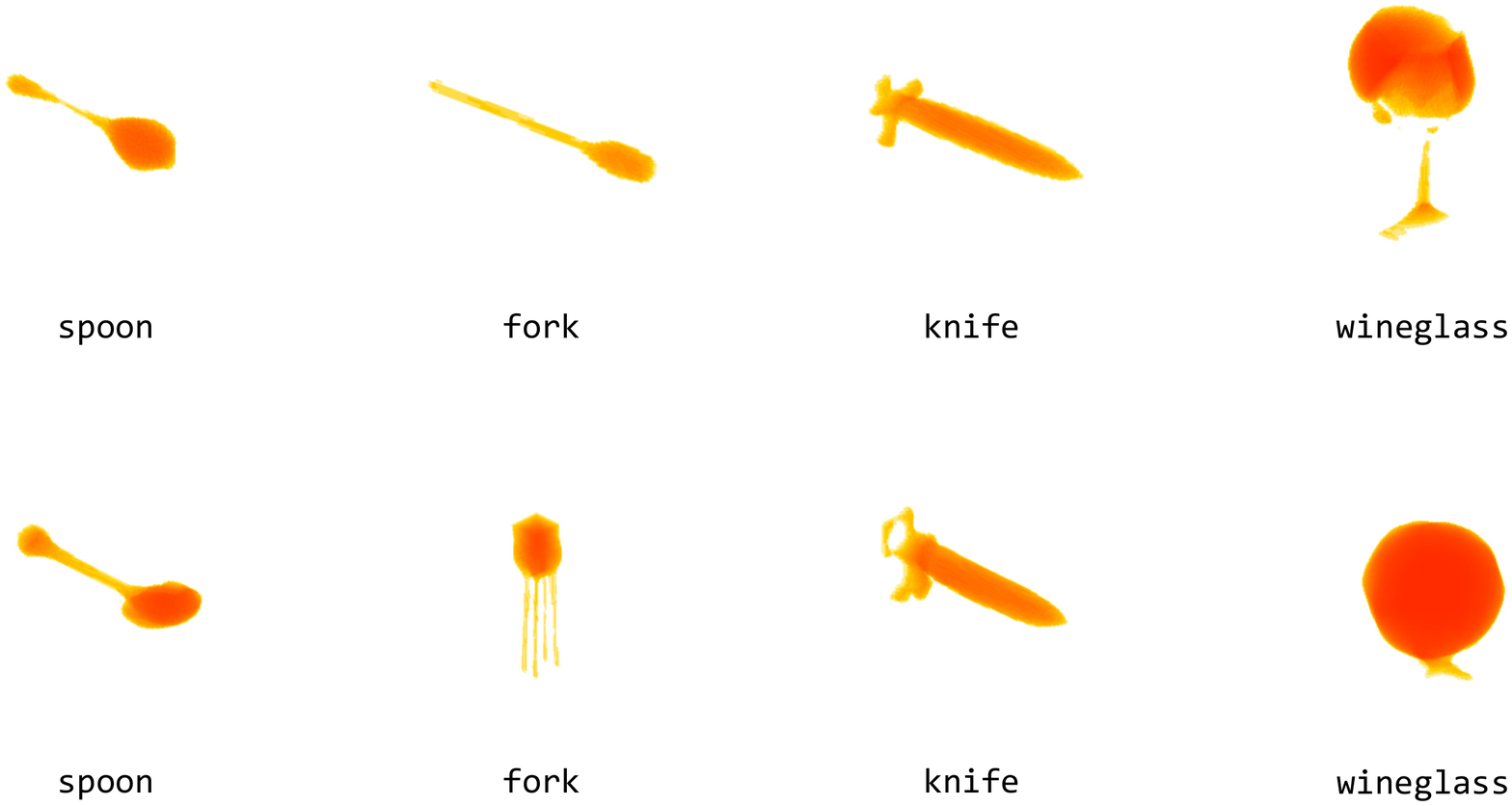} 
\end{tabular}
\caption{Volume renderings for shapes shown in \figref{tab:contrast_ablation}. The top row corresponds to $\lambda_c = 0.01$ and $\tau = 30$, the second row corresponds to $\lambda_c = 0.1$ and $\tau = 30$, the third row corresponds to $\lambda_c = 0.01$ and $\tau = 50$, and the fourth row corresponds to $\lambda_c = 0.1$ and $\tau = 50$.}
\label{tab:fig6_vol_render}
\end{figure}

\clearpage

\bibliography{references}

\end{document}